\documentclass[journal=jacsat,manuscript=article]{achemso}
\usepackage{float}
\usepackage[version=3]{mhchem} % Formula subscripts using \ce{}
\usepackage{booktabs}
\usepackage{multirow}
\usepackage{array}
\usepackage{hyperref}
\usepackage{pifont} % For \cmark and \xmark
\author{Houxu Chen}
\affiliation[CMU-BME]
{Biomedical Engineering, Carnegie Mellon University, Pittsburgh, PA 15213, USA}
 
\author{Achuth Chandrasekhar}
\alsoaffiliation[CMU-MECHE]
{Mechanical Engineering, Carnegie Mellon University, Pittsburgh, PA 15213, USA}

\author{Amir Barati Farimani}
\email{barati@cmu.edu}
\affiliation[CMU-MECHE]
{Mechanical Engineering, Carnegie Mellon University, Pittsburgh, PA 15213, USA}
\alsoaffiliation[CMU-BME]
{Biomedical Engineering, Carnegie Mellon University, Pittsburgh, PA 15213, USA}
\alsoaffiliation[CMU-CHE]
{Chemical Engineering, Carnegie Mellon University, Pittsburgh, PA 15213, USA}
\alsoaffiliation[CMU-MSE]
{Materials Science and Engineering, Carnegie Mellon University, Pittsburgh, PA 15213, USA}
\alsoaffiliation[CMU-ML]
{Machine Learning Department, Carnegie Mellon University, Pittsburgh, PA 15213, USA}

\title{
    \textbf{Pepti-Agent: An AI Agent for Peptide Design and Optimization}
}
\abbreviations{MCP, LLM, AI, ESM}
 
\usepackage{siunitx}
\usepackage{subcaption}
\usepackage{graphicx}
\usepackage{bbm}
\usepackage{amsfonts}
\usepackage{url}
\usepackage{listings}
\usepackage{xcolor}
 
\newcommand*\code[1]{\texttt{#1}}
\newcommand*\Sscore{S}
\newcolumntype{L}[1]{>{\raggedright\arraybackslash}p{#1}}
 
\definecolor{cmdbg}{RGB}{233,243,252}      % commands: light blue
\definecolor{cmdaccent}{RGB}{45,110,190}
\definecolor{promptbg}{RGB}{234,247,236}    % prompts: light green
\definecolor{promptaccent}{RGB}{48,138,58}
\definecolor{pathbg}{RGB}{253,244,228}      % file paths: light amber
\definecolor{pathaccent}{RGB}{205,130,25}
 
\lstdefinestyle{cmd}{backgroundcolor=\color{cmdbg},rulecolor=\color{cmdaccent}}
\lstdefinestyle{path}{backgroundcolor=\color{pathbg},rulecolor=\color{pathaccent}}
 
\definecolor{promptbg}{RGB}{234,247,236}   \definecolor{promptaccent}{RGB}{48,138,58}    % green  – system prompt
\definecolor{userbg}{RGB}{238,240,244}     \definecolor{useraccent}{RGB}{84,96,112}      % slate  – user template
\definecolor{genbg}{RGB}{228,244,245}      \definecolor{genaccent}{RGB}{26,138,148}      % teal   – peptide_generation
\definecolor{predbg}{RGB}{238,236,250}     \definecolor{predaccent}{RGB}{96,79,190}      % violet – property_prediction
\definecolor{refbg}{RGB}{249,236,242}      \definecolor{refaccent}{RGB}{178,72,112}      % rose   – iterative_refinement
 
\usepackage{tcolorbox}
\tcbuselibrary{listings,skins,breakable}
\tcbset{
  promptbase/.style={
    breakable, enhanced, boxrule=0.6pt, arc=1.5pt,
    left=7pt, right=7pt, top=6pt, bottom=6pt,
    listing only,
    listing options={basicstyle=\ttfamily\footnotesize, breaklines=true,
      breakatwhitespace=false, columns=fullflexible, keepspaces=true,
      showstringspaces=false}
  }
}
\newtcblisting{promptbox}{promptbase, colback=promptbg, colframe=promptaccent}
\newtcblisting{userpromptbox}{promptbase, colback=userbg, colframe=useraccent}
\newtcblisting{genpromptbox}{promptbase, colback=genbg, colframe=genaccent}
\newtcblisting{predpromptbox}{promptbase, colback=predbg, colframe=predaccent}
\newtcblisting{refpromptbox}{promptbase, colback=refbg, colframe=refaccent}
\newcommand{\promptlabel}[1]{%
    \par\addvspace{6pt}\noindent\textbf{\footnotesize #1}\par\nobreak\vspace{2pt}}

\begin{document}
 
%%%%%%%%%%%%%%%%%%%%%%%%%%%%%%%%%%%%%%%%%%%%%%%%%%%%%%%%%%%%%%%%%%%%%
%% The "tocentry" environment can be used to create an entry for the
%% graphical table of contents. It is given here as some journals
%% require that it is printed as part of the abstract page. It will
%% be automatically moved as appropriate.
%%%%%%%%%%%%%%%%%%%%%%%%%%%%%%%%%%%%%%%%%%%%%%%%%%%%%%%%%%%%%%%%%%%%%
\begin{tocentry}
\centering
\includegraphics[width=0.96\linewidth]{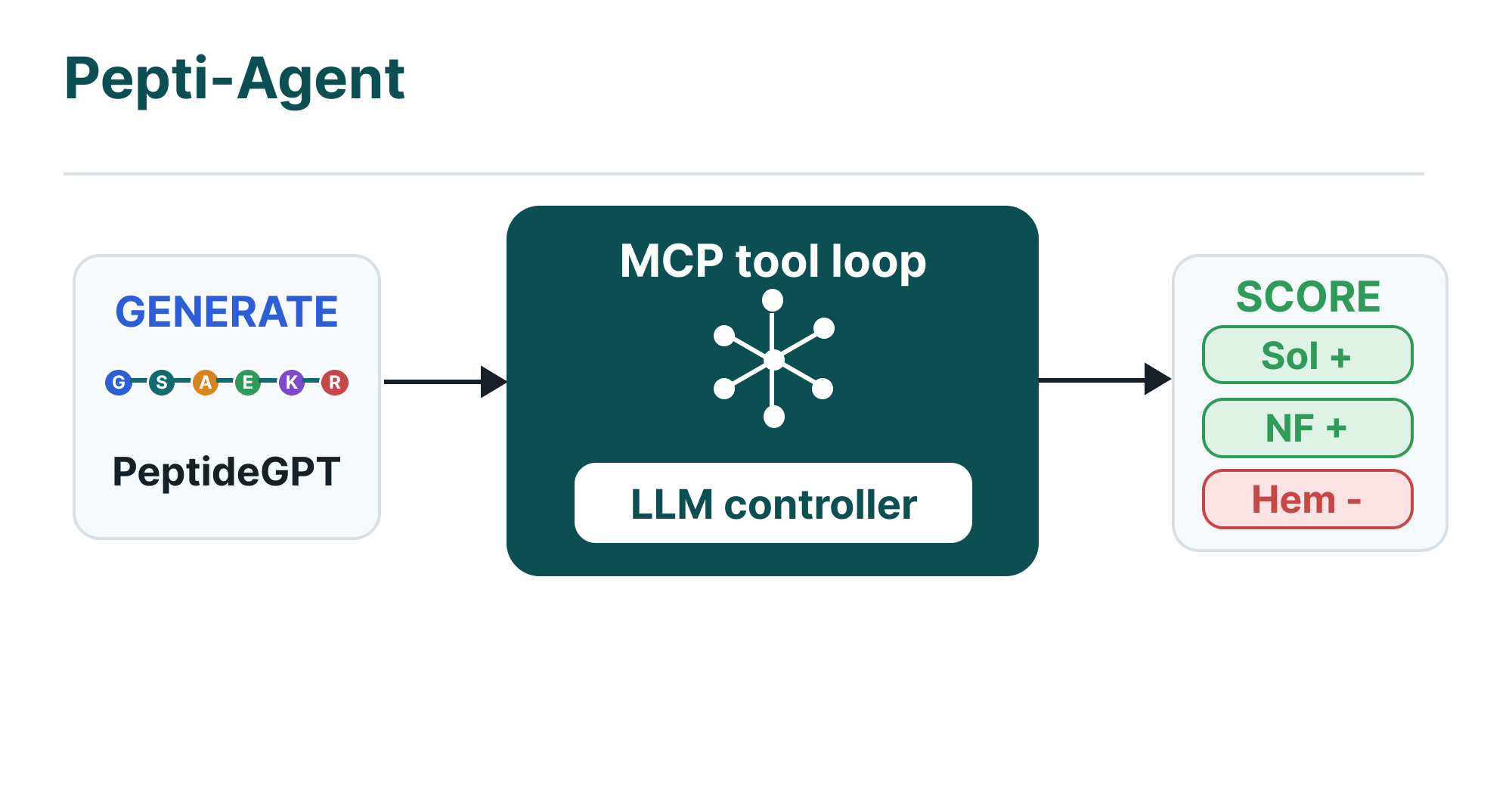}
\end{tocentry}

%%%%%%%%%%%%%%%%%%%%%%%%%%%%%%%%%%%%%%%%%%%%%%%%%%%%%%%%%%%%%%%%%%%%%
%% The abstract environment will automatically gobble the contents
%% if an abstract is not used by the target journal.
%%%%%%%%%%%%%%%%%%%%%%%%%%%%%%%%%%%%%%%%%%%%%%%%%%%%%%%%%%%%%%%%%%%%%
\begin{abstract}
Therapeutic peptides occupy a valuable design space between small molecules and biologics, but their development requires satisfying several competing constraints at once: solubility, hemolytic activity, and nonspecific surface fouling are governed by overlapping sequence features, so improving one property often degrades another \cite{muttenthaler2021peptide,wang2022therapeuticpeptides,luo2023peptidedrugs,julke2025peptidetherapeutics,liu2025peptideproteintherapeutics,peptidebert2023}.
Computational design addresses this by pairing generative models with sequence-based property predictors, iteratively proposing and refining candidates \cite{peptidegpt2024,multipeptide2025,murakami2023multiobjective,peptune2025}.
However, these components are typically wired together as monolithic scripts that are difficult to inspect, extend, or reuse, and they often refine sequences by natural-language reasoning rather than by tracking the evolving multi-property state of each candidate.
We present Pepti-Agent, a closed-loop, peptide-specific framework that exposes generation, property prediction, and single-residue mutation as independently inspectable Model Context Protocol (MCP) tools. A large language model controller invokes these tools and consults live predictor output between calls, so refinement is guided by each sequence's current property profile rather than by language reasoning alone \cite{peptidebert2023,prottrans2022,esmfold2023,mcp2024,react2023,bran2024chemtools,boiko2023autonomous}.
Task-specific PeptideGPT models generate candidates, ProtBERT-based classifiers score solubility, hemolysis, and non-fouling, and two interchangeable mutation operators propose sequence edits. By recording a per-step trace of controller decisions, predictor outputs, and accepted mutations, Pepti-Agent offers a reproducible substrate for benchmarking multi-objective design strategies and for prioritizing candidates for experimental validation.

\end{abstract}
\section{Introduction}
\label{sec:introduction}
 
Peptide therapeutics occupy an important design space between small molecules and large biologics, with advantages in target specificity, potency, and chemical tunability. However, their clinical translation is often limited by development liabilities such as poor stability, low bioavailability, membrane interactions, hemolysis, and nonspecific fouling \cite{muttenthaler2021peptide,wang2022therapeuticpeptides,luo2023peptidedrugs,julke2025peptidetherapeutics,liu2025peptideproteintherapeutics,vrbnjak2024recentadvances}. Useful candidates must therefore balance several properties at once. Peptide design is a constrained multi-objective optimization problem, in which candidates are iteratively proposed, evaluated, and refined \cite{happenn2020,peptidebert2023,multipeptide2025,murakami2023multiobjective,peptune2025}. The present work targets a subset of these liabilities, namely, solubility, hemolysis, and nonspecific fouling, for which trained sequence-based predictors are available; stability, bioavailability, and membrane interactions remain outside the current loop and are discussed in the Limitations subsection.
 
Existing computational approaches commonly pair generative models with predictive models \cite{chen2024aipeptides,wan2024ampml}. Peptide-specific language models such as PeptideGPT enable controllable sequence generation, whereas transformer-based predictors estimate biophysical properties from sequence or combined sequence-structure representations \cite{peptidegpt2024,peptidebert2023,multipeptide2025}. Recent multi-objective peptide design methods further show the value of search-guided optimization when desirable properties conflict \cite{murakami2023multiobjective,peptune2025}. However, these components are often connected through monolithic scripts that are difficult to inspect, extend, or reuse across interfaces. Scientific agents augmented with tools offer an alternative, in which a large language model coordinates external computational tools rather than serving as a stand-alone predictor \cite{react2023,schick2023toolformer,aiscientist2024,boiko2023autonomous,bran2024chemtools,cactus2024,scitoolagent2025,drugpilot2025}.
 
Compared to chemistry-oriented tool agents such as ChemCrow \cite{bran2024chemtools} and broader scientific tool routers such as SciToolAgent \cite{scitoolagent2025} and DrugPilot \cite{drugpilot2025}, Pepti-Agent is designed specifically for optimizing peptide development. Rather than treating peptide design as a general drug-design routing problem, the workflow couples peptide generation, mutation, property prediction, and score aggregation around three practical constraints: solubility, hemolytic activity, and nonspecific surface fouling. The controller keeps track of the predicted property profile during refinement and uses the current predictor output to decide which operation to call next. In this way, the refinement process is guided by the evolving multi-property state of each sequence, rather than by natural-language reasoning alone. The implementation is exposed through the Model Context Protocol (MCP), so the generation, prediction, mutation, and aggregation modules can be inspected and reused from different front ends without changing the underlying model code.
 
We evaluate this peptide-specific workflow through three analyses: feasibility, exhaustive single-substitution search (ES), and trade-offs. We also use the MCP implementation to illustrate the modular tool-integration design. This paper makes four contributions. First, we introduce a closed-loop, MCP-orchestrated agent that integrates PeptideGPT generators, ProtBERT-based classifiers for solubility, hemolysis, and non-fouling, and complementary LLM-guided and ESM2-based mutation operators behind a single inspectable tool interface. Second, a pooled before-and-after analysis of 300 generated peptides shows that the agent improves predicted feasibility while exposing cross-property trade-offs. Third, a matched fixed-length analysis shows that ES reveals the headroom left by the current conservative controller. Fourth, an aggressive-refinement analysis shows that higher matched-start scores become reachable once the agent is allowed to expand its edit policy beyond fixed-length single substitutions.
 
\section{Methods}
\label{sec:methods}
 
\subsection{Agentic Workflow}
\begin{figure}[H]
    \centering
    \includegraphics[width=0.95\linewidth]{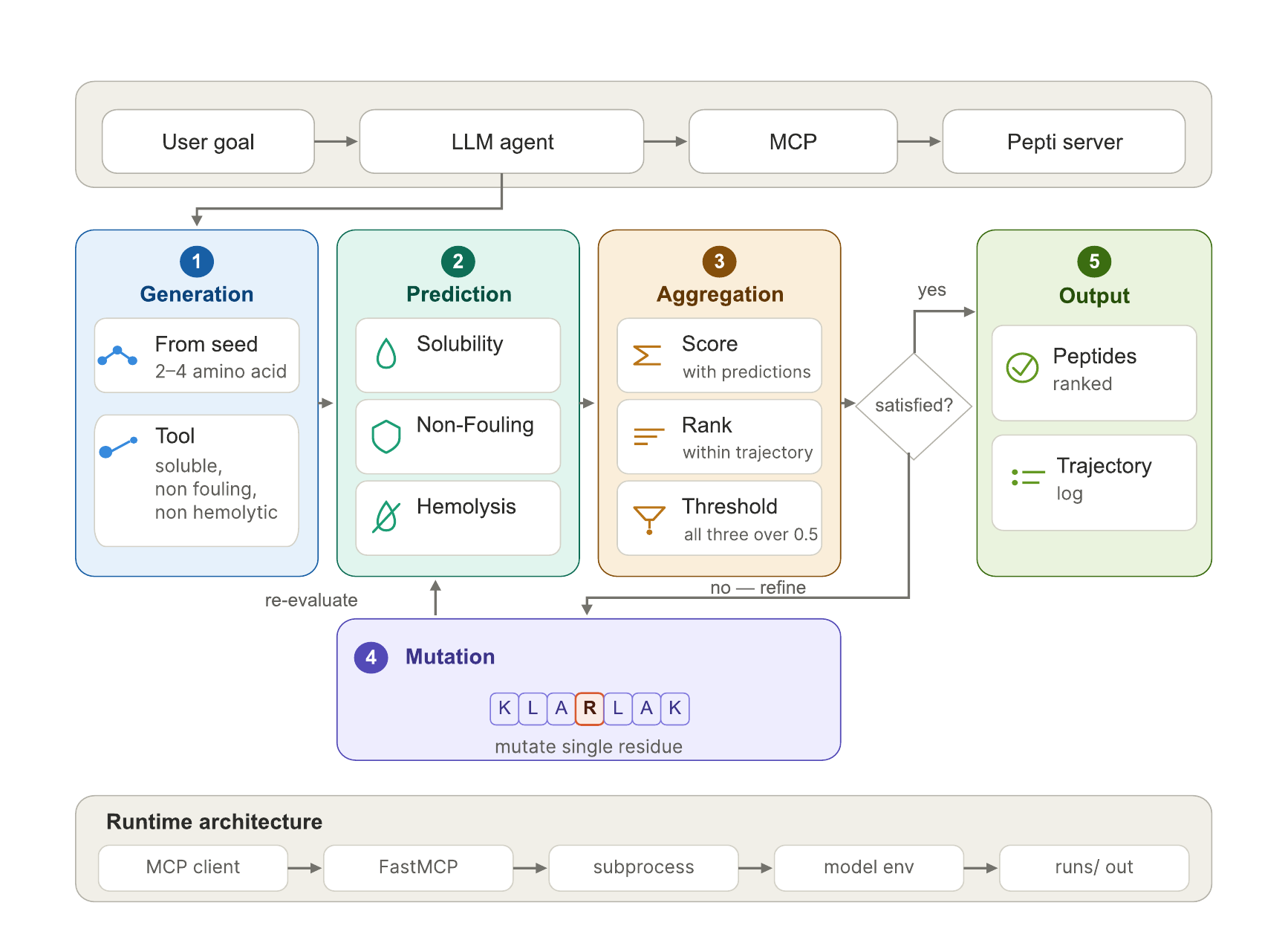}
    \caption{Iterative tool-calling workflow for peptide optimization.
    The agent orchestrates sequence generation, property prediction,
    and mutation-based refinement through a closed-loop optimization process.}
    \label{fig:agent_workflow}
\end{figure}
Pepti-Agent implements a closed-loop optimization workflow (Figure~\ref{fig:agent_workflow}). When the user provides an initial peptide sequence, the system first evaluates that sequence with all property predictors. The resulting solubility, hemolysis, and non-fouling scores define the baseline state. If all target constraints are already satisfied, the workflow terminates immediately. Otherwise, the sequence enters iterative refinement. The loop terminates when (i) all three property thresholds are satisfied, (ii) the configured per-trajectory round budget is reached, or (iii) the trajectory log indicates an oscillation, defined as a proposal that exactly reintroduces a sequence already present in the current trajectory. The current implementation does not impose a wall-clock timeout beyond the round budget.
When no initial peptide is provided, the agent first generates seed candidates using one of the task-specific generators. Each candidate is then evaluated by the property predictors, and the best-performing sequence is selected for further improvement. Refinement proceeds through a series of local edits. At each round, the agent identifies a target property, proposes a targeted mutation, re-evaluates the modified sequence, and records the updated multi-property profile. If the primary mutator stalls or proposes an oscillatory reversal, control can be routed to the ESM2-based fallback mutator.
 
This workflow follows a general act-observe-adapt pattern that resembles tool-using reasoning frameworks such as ReAct and Toolformer, while remaining specialized for peptide optimization \cite{react2023,schick2023toolformer}. It also reflects the design logic of recent chemistry and scientific agents, where tool calls are selected, executed, and inspected as part of an iterative workflow \cite{boiko2023autonomous,bran2024chemtools,cactus2024}. The implementation explicitly avoids one-shot optimization. Every intervention is numerically evaluated before the next tool call is chosen. In the current implementation, however, the loop is optimized for inspectability rather than strict monotonic improvement, in the spirit of self-reflective agent frameworks that retain failed proposals as feedback rather than discarding them \cite{shinn2023reflexion}. Fallback mutation is triggered when the primary mutator stalls, and regressing proposals remain visible in the trajectory log instead of being silently discarded. As a result, the controller exposes its own successes and failure modes directly. This visibility is useful for diagnosis but motivates stronger acceptance and stopping criteria for fully autonomous optimization.
 
\subsection{Model}
 
Pepti-Agent uses a multi-component modeling stack rather than a single end-to-end model. Candidate peptides are first generated by task-specific sequence generators based on PeptideGPT, a generative transformer model for peptides with hemolytic, non-hemolytic, solubility, and non-fouling generation tasks \cite{peptidegpt2024}. Separate fine-tuned checkpoints are used for soluble, non-hemolytic, and non-fouling peptides. Checkpoint identifiers are listed in Section~S2 of the Supporting Information. These generators accept short amino-acid seeds, typically 2 to 4 residues, and stochastically extend them into complete peptide sequences through temperature-controlled top-$p$ sampling \cite{holtzman2020nucleus}. The sequence length and number of returned candidates are exposed as tool arguments.
 
Peptide properties are then estimated using PeptideBERT-style classifiers built on a frozen ProtBERT backbone followed by a shallow linear classification head. This head maps the pooled sequence representation to a single property probability \cite{peptidebert2023,prottrans2022}. Three checkpoints, one each for solubility, hemolysis, and non-fouling, are evaluated independently. This design follows sequence-based peptide-property prediction work, while remaining compatible with later multimodal extensions that combine sequence and structural representations \cite{happenn2020,peptidebert2023,multipeptide2025}. Batched inference over candidate sets produces comparable probabilistic outputs for the agent loop.
 
The refinement stage contains two complementary mutation operators. The primary operator is an LLM-guided single-site mutator that proposes one amino-acid substitution conditioned on the sequence and its current property scores. The prompt template instructs the LLM to (i) preserve the sequence length, (ii) improve the selected target property, and (iii) perturb the other two properties as little as possible. It returns a structured \texttt{(position, new\_residue)} tuple with a short rationale that is stored in the trajectory log. A proposal is accepted when it is syntactically valid (a single substitution in a legal position with one of the 20 canonical amino acids) and does not regress the target property below the value of the previous round; if either condition fails, control is routed to a fallback operator based on the ESM2 masked-language model \cite{rives2021biological,meier2021language} that masks the same position and selects the highest-likelihood non-identity amino acid under its masked-language-model head. This fallback is property-blind by construction. The exact numerical acceptance threshold applied in the current conservative controller is a configurable parameter of the codebase rather than a fixed value of the framework.
 
Together, these components define a modular search loop over peptide sequence space that combines generation, scoring, and two complementary modes of local refinement.
 
\subsection{MCP Framework}
 
Pepti-Agent exposes its core functionality through the Model Context Protocol using a \texttt{FastMCP} server \cite{mcp2024}. Within this framework, each major peptide-engineering capability is registered as a tool with a defined interface. These capabilities include generation tools, property-prediction tools, mutation tools, and a result-aggregation tool. The same layer also exposes prompt resources that describe recommended usage patterns for external agents. This design follows the broader move from stand-alone models toward tool interfaces that make scientific workflows inspectable, composable, and reusable \cite{react2023,bran2024chemtools,boiko2023autonomous,cactus2024,scitoolagent2025}.
 
This MCP abstraction serves two purposes. First, it standardizes access to the peptide-engineering components so that they can be consumed by different front ends without changing the underlying model implementations. Second, it separates low-level computational primitives from high-level orchestration logic. The generation and prediction modules remain domain-specific, while the agent determines how they should be combined during optimization. This separation is consistent with recent scientific agents in which the language model functions as a controller over external capabilities rather than replacing domain tools \cite{bran2024chemtools,boiko2023autonomous,cactus2024,scitoolagent2025}.
 
The implementation supports multiple operating modes, including standard input/output transport, Server-Sent Events transport, and an interactive command-line interface. This flexibility allows the same codebase to function as a local research assistant, a tool server for agent frameworks, or a remotely accessible service. In this sense, MCP acts as the integration layer that transforms a set of peptide-modeling scripts into a composable agent environment.

\subsection{Agent Toolset}
 
The toolset is organized into four categories that cross-reference the components introduced in the Methods section: sequence generation (one tool each for soluble, non-hemolytic, and non-fouling candidates, each wrapping a task-specific PeptideGPT checkpoint), property prediction (one tool each for hemolysis, solubility, and non-fouling, using the matched PeptideBERT-style classifiers), mutation (the LLM-guided single-site mutator and the ESM2-based fallback mutator), and result aggregation. Table~\ref{tab:mcp_tools} gives the concrete inventory and tool signatures.
 
The prediction tools accept raw sequences, delimited lists, JSON payloads, and file paths, which lowers integration overhead across workflows. The aggregation tool computes derived labels indicating whether a sequence simultaneously satisfies the non-hemolytic, soluble, and non-fouling thresholds, and reports a composite developability score that favors high solubility, high non-fouling probability, and low hemolysis probability. We use the linear composite
\begin{equation}
S(\mathbf{x}) = p_{\mathrm{sol}}(\mathbf{x}) + p_{\mathrm{nf}}(\mathbf{x}) + \bigl(1 - p_{\mathrm{hem}}(\mathbf{x})\bigr),
\label{eq:composite}
\end{equation}
where $p_{\mathrm{sol}}$, $p_{\mathrm{nf}}$, and $p_{\mathrm{hem}}$ are the predictor probabilities for solubility, non-fouling, and hemolysis. $S$ lies in $[0,3]$ and is used both to rank candidates within a trajectory and as the primary scalar outcome compared across methods in the Results section, which refers to this definition throughout. We keep the weighting between objectives uniform, following scalarization approaches in multi-objective molecular design \cite{fromer2023computer}, so that the ranking stays inspectable. Richer Pareto-based ranking is a methodological extension, discussed in the algorithm-improvement paragraph of the Future Work subsection.
 
Pepti-Agent also ships prompt definitions that tell an autonomous agent how to use the toolset. These serve as both documentation and behavioral constraints. They specify, for instance, that every property predictor is called after each mutation and that optimization continues until a valid stopping condition is reached. The framework thus pairs low-level computational primitives with an explicit controller policy, as in recent multi-tool scientific-agent systems \cite{cactus2024,scitoolagent2025,drugpilot2025}. Unlike broader tool routers, the policy here is restricted to a peptide-specific design loop and is evaluated only through predictor-defined peptide properties.

\begin{table}[H]
    \centering
    \small
    \caption{MCP tools exposed by the Pepti-Agent server. Categories: generation (Gen.), prediction (Pred.), mutation (Mut.), aggregation (Agg.). The \texttt{generate\_*} and \texttt{predict\_*} tools each share a signature within their group and are listed once.}
    \label{tab:mcp_tools}
    \begin{tabular}{>{\raggedright\arraybackslash}p{0.34\linewidth} p{0.07\linewidth} >{\raggedright\arraybackslash}p{0.23\linewidth} >{\raggedright\arraybackslash}p{0.24\linewidth}}
        \toprule
        Tool name & Cat. & Input & Returns \\
        \midrule
        \texttt{generate\_\allowbreak non\_\allowbreak hemolytic\_\allowbreak peptide}, \texttt{generate\_\allowbreak soluble\_\allowbreak peptide}, \texttt{generate\_\allowbreak non\_\allowbreak fouling\_\allowbreak peptide} & Gen. & optional \texttt{seed} & Generated peptide sequence (see caption) \\
        \addlinespace
        \texttt{predict\_\allowbreak hemolysis}, \texttt{predict\_\allowbreak solubility}, \texttt{predict\_\allowbreak non\_\allowbreak fouling} & Pred. & \texttt{sequences} & Per-sequence $p_{\mathrm{hem}}$, $p_{\mathrm{sol}}$, $p_{\mathrm{nf}}$ scores respectively (see caption) \\
        \addlinespace
        \texttt{llm\_\allowbreak mutate\_\allowbreak sequence} & Mut. & \texttt{sequence}, three property scores, \texttt{target\_property} & Same-length mutated sequence with position, residue change, and rationale \\
        \addlinespace
        \texttt{esm2\_\allowbreak mutate\_\allowbreak sequence} & Mut. & \texttt{sequence} & Same-length mutated sequence selected by ESM2 masked-LM likelihood \\
        \addlinespace
        \texttt{evaluate\_\allowbreak results} & Agg. & Three JSON prediction outputs & Unified per-sequence summary used for ranking and stopping \\
        \bottomrule
    \end{tabular}
\end{table}
 
Each tool in Table~\ref{tab:mcp_tools} is registered through the \texttt{@mcp.tool()} decorator in the server entry point and is callable identically from the interactive CLI, from external MCP clients over standard input/output transport, or from Server-Sent Events clients on a configurable port. The server also exposes three MCP prompts, \texttt{peptide\_generation}, \texttt{property\_prediction}, and \texttt{iterative\_refinement}, encoding the usage patterns above. The full prompt text is reproduced in Section~S3 of the Supporting Information. The implementation is Python on \texttt{torch}, \texttt{transformers}, \texttt{fair-esm}, \texttt{mcp}, and \texttt{openai} \cite{prottrans2022,esmfold2023,mcp2024}, selects CUDA when available and falls back to CPU, and resolves Hugging Face models from a local cache to avoid repeated downloads. The companion repository, linked in the Supporting Information, contains the full prompt text, the checkpoint identifiers, the module layout, and the CLI driver for the refinement protocol described next.
 
The implementation is written in Python and relies on \texttt{torch}, \texttt{transformers}, \texttt{fair-esm}, \texttt{mcp}, and \texttt{openai} \cite{prottrans2022,esmfold2023,mcp2024}. The code is device-aware and selects CUDA when available, otherwise falling back to CPU execution. Hugging Face model resolution is handled locally through cache inspection, reducing repeated downloads when models are already available on disk.
 
The codebase is compact and modular. The top-level server file defines the MCP server, command-line behavior, and tool registration. The \texttt{tools} directory contains implementations for generation, prediction, mutation, and structured result aggregation. The \texttt{configs} directory defines the ProtBERT-style prediction wrapper, and the \texttt{prompts} directory contains reusable instruction templates for generation, prediction, and refinement. The codebase includes all components required to run the system and inspect its behavior.
 
\subsection{Evaluation Protocol}
\label{sec:evaluation}
 
The Results section uses two evaluation regimes that share the same predictors and composite score from Equation~\ref{eq:composite}. Detailed methods are given in Section~S4 of the Supporting Information. The first is a pooled refinement evaluation on 300 generated candidate peptides (first Results subsection). The pool is constructed by pairing each candidate produced by the soluble, non-hemolytic, and non-fouling PeptideGPT generators with its counterpart produced by the conservative agent under a single fixed prompt template and configuration. Each candidate is refined under the same per-trajectory round budget and stopping criteria described in the Methods section, and the same random seed is used across the pool to standardize generator and mutator sampling.
The second regime is a matched-start analysis (the same starts are reused across all settings) on 24 representative starting peptides (second and third Results subsections). The 24 starts are drawn from the same pool by stratifying each of the three generator-source pools (soluble, non-fouling, and non-hemolytic; the non-hemolytic pool is abbreviated ``Hem'' in the figures) into composite-score quartiles and selecting two peptides per quartile, for a total of $2 \times 4 \times 3 = 24$ starts. Within each quartile, the two starts are selected by a deterministic, score-ordered rule so that the same 24 starts are reused across the conservative-agent, ES, and aggressive-agent comparisons.
The conservative-agent, aggressive-agent, and ES outputs reported in the second and third Results subsections are evaluated against the same 24 starts; the conservative baseline used in the aggressive-vs-conservative comparison is the identical conservative agent used in the ES analysis. A formal benchmark harness and automated test suite are deferred to future releases (see Future Work).
 
%%\clearpage
\section{Results}
\label{sec:results}
 
\noindent This section is organized as a staged evaluation of Pepti-Agent. We first ask whether conservative-agent refinement improves predicted feasibility across a pooled set of generated peptides. We then use ES as a local upper bound, defined as the highest composite score reachable, to quantify how much optimization headroom remains within the same edit-depth regime. Finally, we evaluate an aggressive refinement setting to test whether broader edit actions can reach higher composite scores. This organization separates three claims: feasibility improvement, local-search headroom, and action-space expansion (edit types beyond fixed-length single substitutions, such as insertions, deletions, and length changes).
 
\subsection{Conservative Refinement and Feasibility}
\label{sec:pooled_refinement}
 
We first evaluated whether conservative Pepti-Agent refinement improves the property profile of generated peptides when all generator origins are pooled. The dataset contained 300 paired candidates, with each pair consisting of an original generated peptide and its refined counterpart. Generator labels were used only to organize files and were not treated as biological groups in the analysis.
 
\begin{figure}[H]
    \centering
    \includegraphics[width=\linewidth]{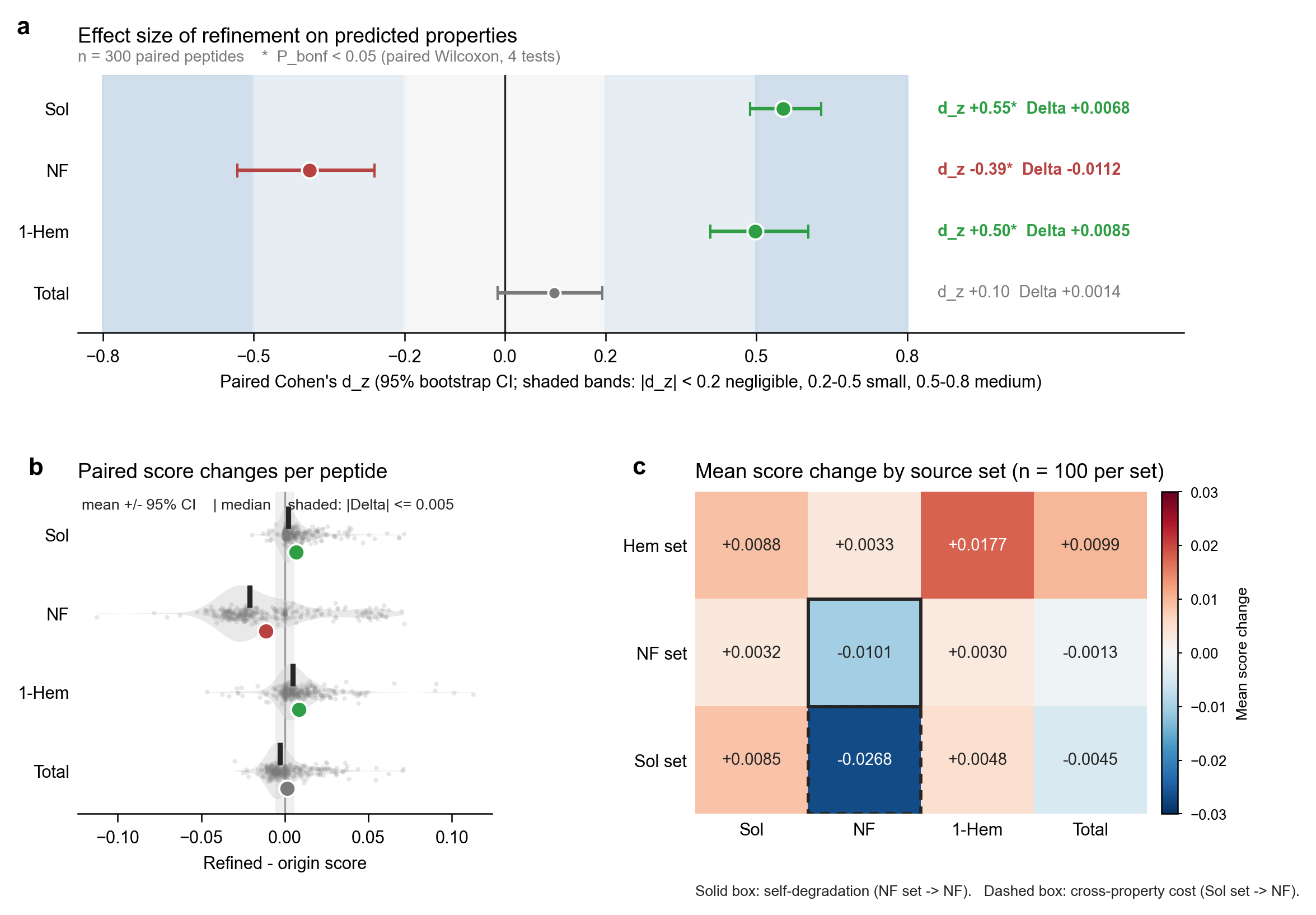}
    \caption{Pooled paired refinement, 300 peptides. (a) Refinement effect size per property and composite $S$ (paired Cohen's $d_z$, 95\% bootstrap CIs); bands mark negligible/small/medium ranges, $\Delta$ annotated. (b) Per-peptide changes (refined $-$ origin) per property and $S$; band marks $|\Delta|<0.005$. (c) Mean change by source set ($n=100$: Sol, NF, Hem), rows source pool, columns property; solid box, non-fouling self-degradation; dashed box, solubility cross-property cost. Statistics in text.}
    \label{fig:pooled_refinement}
\end{figure}

At the feasibility level, the number of peptides failing any property threshold fell from 2 of 300 before refinement to 0 of 300 after refinement (Figure~\ref{fig:pooled_refinement}). Because only 2 of 300 candidates failed the $0.5$ threshold at baseline, this feasibility-recovery endpoint is intentionally lenient and should be read as ``no candidate is left below threshold'' rather than as a strong demonstration of optimization performance; we return to this threshold-sensitivity caveat in the Limitations subsection. On the predicted objectives, mean predicted solubility shifted from 0.717 to 0.724 and mean predicted hemolysis from 0.390 to 0.381, while mean predicted non-fouling decreased from 0.649 to 0.637; every refined peptide nonetheless remained above the non-fouling reporting threshold of 0.5. The corresponding mean composite score $S$ from Equation~\ref{eq:composite} shifted by an amount comparable to the per-property mean shifts, consistent with a small net improvement coupled to a one-property trade-off.
 
Paired Wilcoxon signed-rank tests with Bonferroni correction across four tests (Figure~\ref{fig:pooled_refinement}a) support these directional shifts: predicted solubility and the desired-direction hemolysis measure $1-p_{\mathrm{hem}}$ each show a significant medium-sized positive paired Cohen's $d_z$ \cite{lakens2013effect} ($+0.55$ and $+0.50$ respectively), while predicted non-fouling shows a significant small-to-medium negative $d_z$ ($-0.39$). The mean composite-score effect is small and not significant ($d_z=+0.10$), consistent with the per-property gains and the non-fouling cost canceling at the aggregate level. Disaggregating by source set (Figure~\ref{fig:pooled_refinement}c) further shows that the composite-score gain is concentrated in peptides drawn from the hemolysis-generator pool, whereas refining peptides drawn from the non-fouling-generator pool reduces their own predicted non-fouling, a self-degradation effect that is consistent with the controller-level limitations discussed below. Taken together, the supported claim from this analysis is trade-off-aware feasibility recovery: conservative Pepti-Agent refinement is able to drive every peptide in the pool above the three reporting thresholds simultaneously, while accepting a statistically significant but small mean decrease in predicted non-fouling.
 
\subsection{Fixed-Length Exhaustive Search Analysis}
\label{sec:fixed_length_oracle}
 
\begin{figure}[H]
    \centering
    \includegraphics[width=\linewidth]{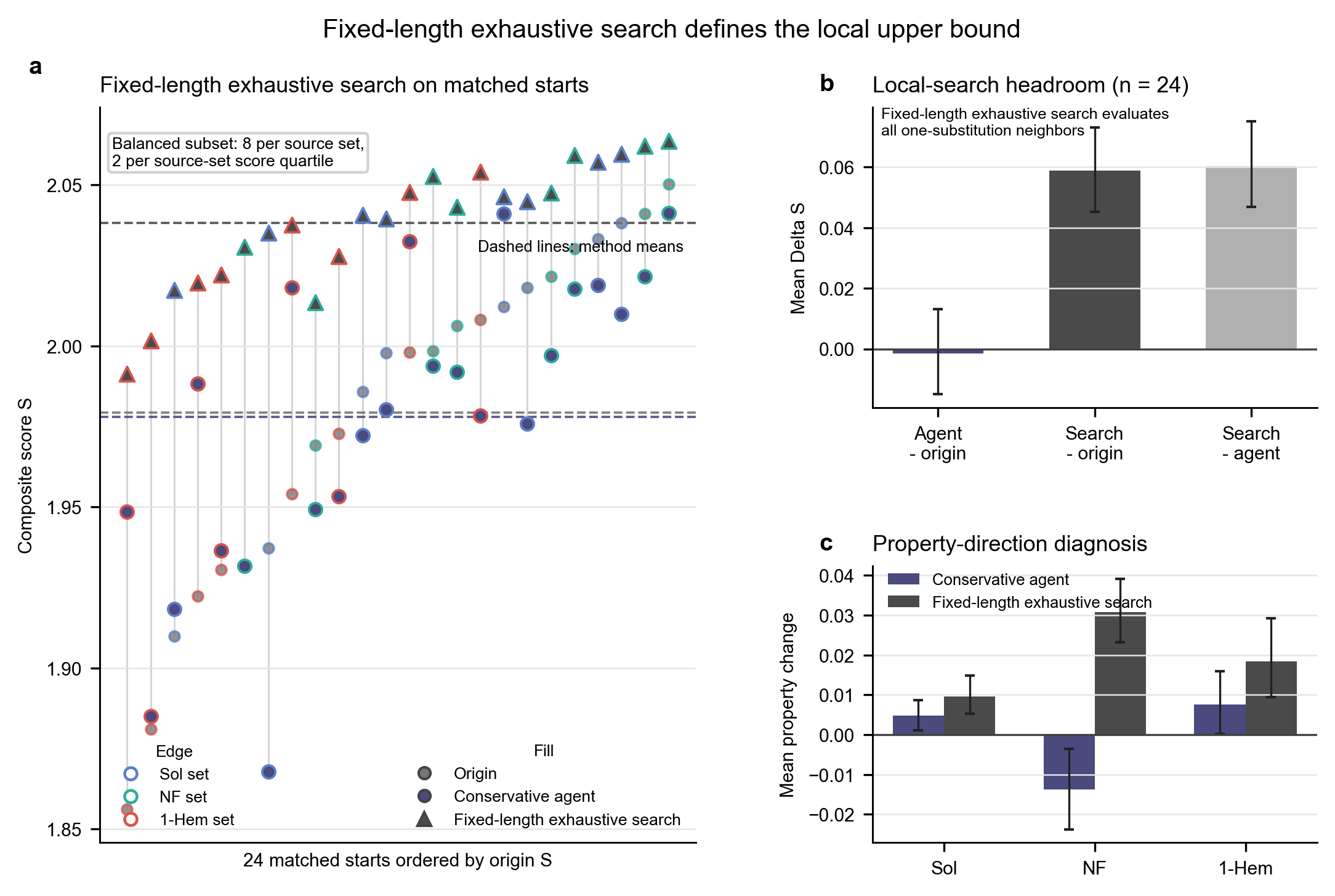}
    \caption{Exhaustive fixed-length search (ES) as local upper bound. (a) Composite $S$ for 24 matched starts, conservative outputs, and best ES; ordered by origin $S$, marker edge denotes source set, dashed lines method means. (b) Headroom ($n=24$): mean $\Delta S$ for agent, ES, and ES vs.\ agent, 95\% bootstrap CIs. (c) Mean desired-direction changes (Sol, NF, $1-p_{\mathrm{hem}}$), conservative vs.\ ES. ES is a same-length upper bound, not compute-matched.}
    \label{fig:fixed_length_oracle}
\end{figure}

We next asked whether the conservative agent reaches the best sequence available within its nominal local edit regime. To answer this, we constructed a matched ES over 24 representative starting peptides: two peptides from each composite-score quartile within each of the soluble, non-fouling, and non-hemolytic source pools. For each start of length $L$, we exhaustively enumerated all possible single-residue substitutions over the 20 canonical amino acids, scored each with the same property predictors used by the agent, and selected the original or mutant sequence with the highest composite score from Equation~\ref{eq:composite}. Because this neighborhood is searched in full rather than sampled, the selected variant is by construction the best score attainable under a single fixed-length substitution, making ES a local upper bound on the conservative controller's own edit regime. For the matched comparison in Figure~\ref{fig:fixed_length_oracle}, the corresponding refined Pepti-Agent peptides were paired with the same origins by the source pool and the original index, while the ES output was paired by the start identifier and the origin sequence.
 
This baseline should not be interpreted as a compute-matched non-agent competitor. It uses the same property predictors and scalar score as the agent comparison, but it exhaustively evaluates the full one-substitution neighborhood and is therefore a local upper bound under the same fixed-length, one-edit search space. However, the comparison is informative because it quantifies the upper bound left by the current conservative controller.
 
ES improved the composite score for all 24 representative starts (Table~\ref{tab:non_agent_control}). Full results tables and source data are in Section~S5 of the Supporting Information. The mean $S$ increased from 1.979 to 2.038, corresponding to a mean gain of 0.059; bootstrap-derived 95\% confidence intervals for this and the subsequent matched-start contrasts are shown in Figure~\ref{fig:fixed_length_oracle}b. The selected sequences also improved all three property directions on average: predicted solubility increased from 0.718 to 0.728, predicted non-fouling increased from 0.654 to 0.685, and predicted hemolysis decreased from 0.393 to 0.375. Because all 24 starting peptides had already satisfied the three reporting thresholds, this result reflects optimization within the feasible region rather than rescue of failing candidates. We report the matched-start contrasts as mean shifts with bootstrap CIs in Figure~\ref{fig:fixed_length_oracle}; given the moderate sample size ($n=24$) and the single-stratification design, these effects should be interpreted as descriptive search-space diagnostics (a description of where each method can reach in sequence space, not a benchmark claim) rather than as inferential effect estimates over an arbitrary peptide population.
 
\begin{table}[H]
    \centering
    \small
    \caption{Exhaustive fixed-length single-substitution search on 24 representative starting peptides. Positive values indicate improvement in the desired direction, and an up arrow ($\uparrow$) on a $\Delta$ column marks that higher is better; $p_{\mathrm{antihem}}=1-p_{\mathrm{hem}}$ reports the desired reduction in predicted hemolysis.}
    \label{tab:non_agent_control}
    \begin{tabular}{lrrrrrrr}
        \toprule
        Source & $n$ & $S_{\mathrm{orig}}$ & $S_{\mathrm{ES}}$ & $\Delta S\,\uparrow$ & $\Delta p_{\mathrm{sol}}\,\uparrow$ & $\Delta p_{\mathrm{nf}}\,\uparrow$ & $\Delta p_{\mathrm{antihem}}\,\uparrow$ \\
        \midrule
        Soluble & 8 & 1.992 & 2.043 & 0.051 & 0.016 & 0.019 & 0.016 \\
        Non-fouling & 8 & 2.006 & 2.047 & 0.041 & 0.004 & 0.026 & 0.010 \\
        Hemolysis & 8 & 1.940 & 2.025 & 0.085 & 0.009 & 0.047 & 0.029 \\
        All & 24 & 1.979 & 2.038 & 0.059 & 0.010 & 0.031 & 0.018 \\
        \bottomrule
    \end{tabular}
\end{table}
 
The mean gap between ES and the agent clarified the interpretation of this result (Figure~\ref{fig:fixed_length_oracle}). On the same 24 starting sequences, conservative Pepti-Agent refinement increased the mean predicted solubility from 0.718 to 0.723 and reduced the mean predicted hemolysis from 0.393 to 0.385, but the mean predicted non-fouling decreased from 0.654 to 0.640, so the mean $S$ decreased slightly from 1.979 to 1.978, with composite-score improvement in only 8 of 24 agent-refined outputs. By contrast, the ES-selected sequence had a higher $S$ than the corresponding conservative-agent output in all 24 cases, with a mean gap of 0.060 in $S$.
 
This 24-start outcome should be read alongside the pooled result in the first Results subsection rather than as a contradiction of it. The pooled analysis is a feasibility statement over a heterogeneous 300-peptide pool that contains threshold-failing candidates: in that regime, the conservative controller is effective precisely because it can move the small set of failing peptides above the reporting thresholds, which dominates the pool-level summary. The 24-start sample, in contrast, is drawn by quartile-stratified selection from peptides that already satisfied the three thresholds, so there are no feasibility failures left to rescue. On that already-feasible subset, performance is determined entirely by within-feasible composite-score optimization, which is where the current conservative policy is weak. The pooled and matched-start results are therefore consistent under a single reading: the conservative controller recovers feasibility but does not yet optimize $S$ within the feasible region. This is a controller-level limitation, not a contradiction; it identifies what the current policy is missing rather than invalidating the framework. The current conservative policy is useful as an inspectable refinement workflow, but it does not yet implement the acceptance and rejection, candidate buffering, or local-neighborhood search needed to compete with exhaustive one-step optimization.
 
\subsection{Aggressive Refinement and Action Space}
\label{sec:aggressive_action_space}
 
We then evaluated whether broader agent actions can access higher-scoring regions of the predictor-defined landscape. The aggressive refinement setting used the same property predictors and composite score but relaxed the conservative edit policy. This comparison is therefore not edit-depth matched (it is not run under the same allowed number and type of edits) to either the conservative agent or ES.
 
\begin{figure}[H]
    \centering
    \includegraphics[width=\linewidth]{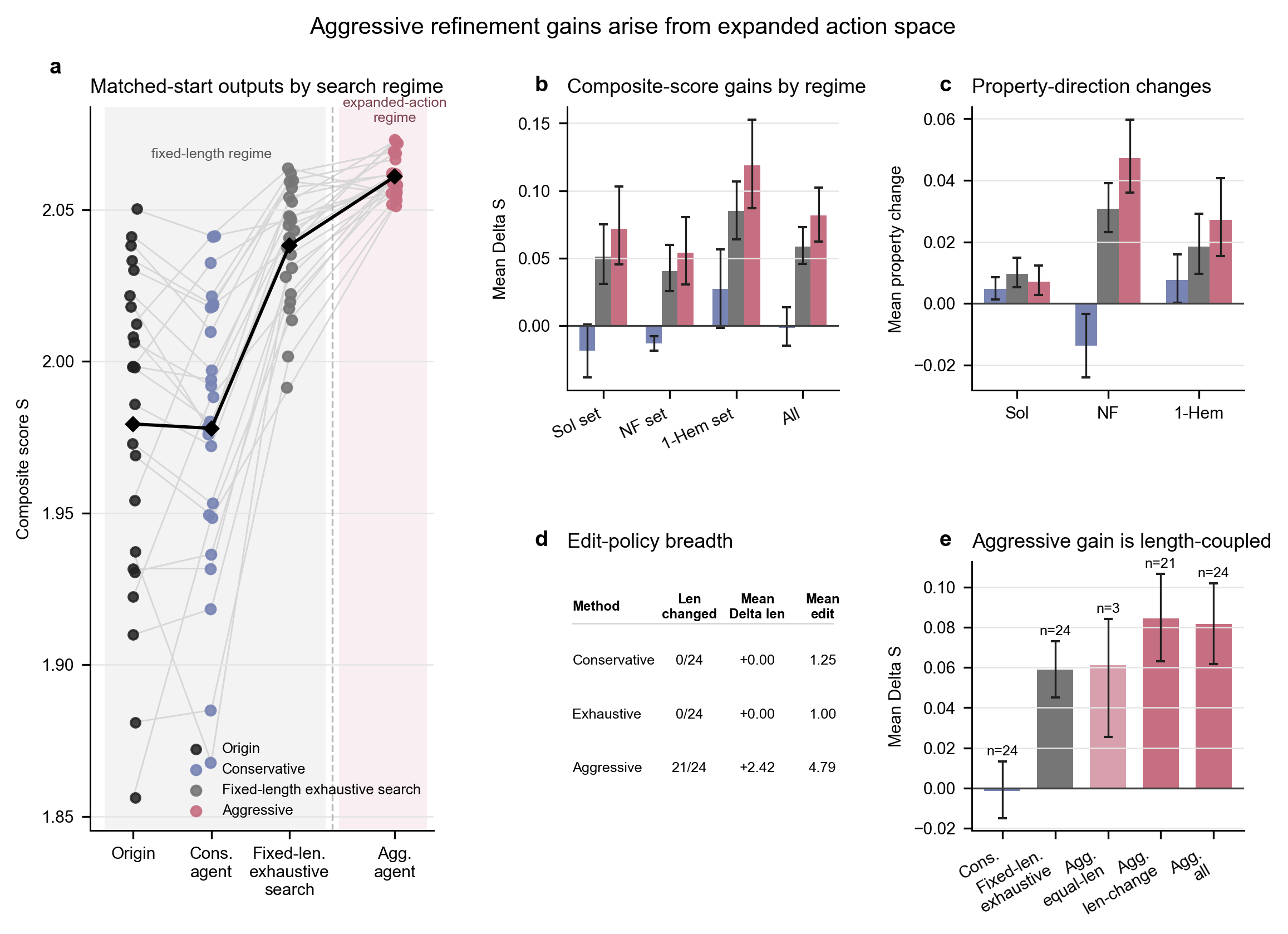}
    \caption{Aggressive gains from expanded action space. (a) Composite scores across four regimes (origin, conservative, ES, aggressive); shading separates fixed-length from expanded-action, grey lines connect starts, heavy line traces means. (b) Mean $\Delta S$ vs.\ origin, ES and aggressive, by source set and overall. (c) Desired-direction changes (Sol, NF, $1-p_{\mathrm{hem}}$), ES vs.\ aggressive. (d) Edit-policy breadth (conservative, ES, aggressive): starts with length change, mean length change, mean edit distance. (e) Aggressive $\Delta S$ by equal-length vs.\ length-changing outputs vs.\ conservative and ES.}
    \label{fig:aggressive_refinement}
\end{figure}

Aggressive refinement improved the composite score for all 24 starts (Figure~\ref{fig:aggressive_refinement}, Table~\ref{tab:aggressive_refinement}). The mean $S$ increased from 1.979 to 2.061, a gain of 0.082 (bootstrap 95\% CI in Figure~\ref{fig:aggressive_refinement}b). The gain was driven primarily by improved predicted non-fouling and the desired-direction hemolysis measure $1-p_{\mathrm{hem}}$: predicted non-fouling increased from 0.654 to 0.701, predicted hemolysis decreased from 0.393 to 0.366, and predicted solubility increased from 0.718 to 0.725. All aggressive-refined candidates satisfied the three reporting thresholds, but, as with ES, this within-feasible improvement is reported descriptively on the same 24 starts and should not be extrapolated beyond them. In head-to-head pairing with the conservative-agent outputs from the ES analysis, using the same 24 conservative-agent trajectories on the same 24 starts with no re-run or reseeding, aggressive refinement produced a higher $S$ in all 24 cases, with a mean advantage of 0.083. It also exceeded the ES output in 22 of 24 cases, with a mean advantage of 0.023. However, this gain was coupled to a substantial change in edit policy: aggressive refinement changed peptide length in 21 of 24 starts, with a mean length increase of 2.42 amino acids. Stratifying by whether output length was retained (Figure~\ref{fig:aggressive_refinement}e) makes this coupling explicit: the 3 aggressive outputs that kept the origin length achieved a mean $\Delta S$ comparable to ES on the same 24 starts, while the 21 aggressive outputs that changed length achieved a larger mean $\Delta S$ that drives the regime-level advantage. Because ES is restricted to single-residue substitutions at constant length, the advantage of aggressive refinement over ES should therefore be interpreted as an expanded-action result rather than as evidence that the agent is a better optimizer under the same search space.
 
How this comparison should be read depends on the precise specification of the aggressive setting, including which fixed-length constraints are relaxed, which additional edit types (single-site insertion, deletion, multi-site substitution, or full re-prompting) are enabled, and whether the LLM prompt template differs from the conservative case.
This distinction is central to the interpretation of the agent. The conservative controller fails to reach the fixed-length local upper bound, indicating that stronger local search and acceptance criteria are needed within the same edit depth. The aggressive controller reaches higher composite scores, but largely by changing the accessible sequence space. We therefore offer the following interpretation as a hypothesis for future testing rather than as an established conclusion: in this predictor-defined peptide-design setting, action-space design appears to be at least as important as the planner itself; isolating these two factors requires the edit-policy-controlled benchmark outlined in the Future Work subsection.
 
\begin{table}[H]
    \centering
    \small
    \caption{Aggressive Pepti-Agent refinement on the same 24 representative starting peptides. Positive values indicate improvement in the desired direction, and an up arrow ($\uparrow$) on a $\Delta$ column marks that higher is better; $p_{\mathrm{antihem}}=1-p_{\mathrm{hem}}$ reports the desired reduction in predicted hemolysis.}
    \label{tab:aggressive_refinement}
    \begin{tabular}{lrrrrrrr}
        \toprule
        Source & $n$ & $S_{\mathrm{orig}}$ & $S_{\mathrm{aggr}}$ & $\Delta S\,\uparrow$ & $\Delta p_{\mathrm{sol}}\,\uparrow$ & $\Delta p_{\mathrm{nf}}\,\uparrow$ & $\Delta p_{\mathrm{antihem}}\,\uparrow$ \\
        \midrule
        Soluble & 8 & 1.992 & 2.063 & 0.072 & 0.014 & 0.032 & 0.026 \\
        Non-fouling & 8 & 2.006 & 2.060 & 0.054 & 0.000 & 0.039 & 0.014 \\
        Hemolysis & 8 & 1.940 & 2.059 & 0.119 & 0.007 & 0.070 & 0.042 \\
        All & 24 & 1.979 & 2.061 & 0.082 & 0.007 & 0.047 & 0.027 \\
        \bottomrule
    \end{tabular}
\end{table}

\section{Discussion}
\label{sec:discussion}
 
The three Results subsections together describe a single, coherent picture of what conservative Pepti-Agent does well, where it falls short, and what the action-space comparison can and cannot say about agentic design.
 
Taken across the three Results subsections, the findings combine into a coherent picture. The pooled-refinement analysis (first Results subsection) establishes that the current conservative controller is effective as a feasibility-recovery procedure: across 300 generated peptides, it brings every remaining threshold-failing candidate above the three reporting thresholds simultaneously, while accepting a small mean decrease in predicted non-fouling. The matched-start ES analysis (second Results subsection) then narrows the lens to a quartile-stratified subset of starts that are already feasible, and shows that within that feasible subset the same conservative controller does not reach the best single-substitution, fixed-length neighbor: ES outscores it in all 24 of 24 cases. The aggressive-refinement analysis (third Results subsection) shows that broader edit policies can move past ES on most matched starts, but do so by changing peptide length in 21 of 24 cases. Together, the three subsections support two claims of limited scope: a capability claim, that the workflow recovers feasibility on heterogeneous pools; and a diagnostic claim, that within-feasible composite-score optimization is bounded both by the controller's acceptance criteria and by the edit policy it is allowed to use.
 
\paragraph{Tool-Agent Context.} The contrast drawn in the Introduction with ChemCrow \cite{bran2024chemtools}, SciToolAgent \cite{scitoolagent2025}, and DrugPilot \cite{drugpilot2025} is exercised in this work at the systems level rather than through a head-to-head benchmark. The peptide-specific toolset is demonstrated by the use of separate soluble, non-hemolytic, and non-fouling generators and matched property predictors throughout the three Results subsections. The trajectory-aware controller is instantiated whenever the controller chooses the next refinement step from live predictor output rather than free-form reasoning, and is what the ES analysis is able to bound from above. The MCP integration layer is realized at the implementation level by exposing every component, including generation, prediction, mutation, and aggregation, as inspectable, reusable tools. None of these three points constitutes evidence of superiority over the cited systems, which target different domains and different evaluation protocols. A direct comparison would require translating the Pepti-Agent tools into the toolset abstraction expected by each of those agents, which is outside the scope of the present systems demonstration and is listed as future work in the Future Work subsection.
 
\paragraph{Controller Gap.} ES is, by construction, the best one-edit candidate within the same predictor-defined landscape; the conservative controller can in principle reach it but is observed not to. Three implementation-level mechanisms are consistent with the matched-start outcome and follow from the workflow described in the Methods section. First, the controller does not maintain a candidate buffer over the local neighborhood: it commits to a single LLM-proposed substitution per round and is therefore biased toward the proposal distribution of the LLM rather than the score distribution over the full neighborhood. Second, the current acceptance rule, as described in the Methods section, gates proposals on per-property non-regression of the targeted property and on syntactic validity, but does not enforce a strict explicit acceptance/rejection criterion on the composite score $S$; a proposal that improves the targeted property while reducing $S$ overall can still be accepted, which is consistent with the observed mean-$S$ decrease on the 24-start subset. Third, the conservative controller exposes oscillatory and regressing proposals in the trajectory log for inspectability rather than silently discarding them, which is appropriate for diagnosis but is exactly what ES does not do. Taken together, the gap to ES is best explained as a missing acceptance/rejection and neighborhood-search layer on top of an otherwise well-behaved tool-calling loop. The Future Work subsection targets each of these mechanisms directly.
 
\section{Limitations and Future Work}
\label{sec:limits}
 
\subsection{Limitations}
\label{sec:limitations_detail}
 
We discuss three main limitations of the current work below; the Future Work subsection describes corresponding directions for addressing them.
 
\paragraph{Predictor Dependence.} The entire optimization loop is gradient-free with respect to ground-truth biology and depends on PeptideBERT-style sequence classifiers. Therefore, current evidence shows movement within a predictor-defined landscape rather than an experimentally confirmed improvement in peptide behavior. The agent can only follow the signals exposed by these predictors, so the apparent dynamic range of each result is bounded by their calibration, training data, and effective resolution. Because no wet-lab hemolysis, solubility, or non-fouling measurements were performed, we cannot yet distinguish true biophysical improvement from model-resolution artifacts.
 
\paragraph{Sequence-Only Scope.} The agent operates entirely on amino-acid sequence and does not inspect three-dimensional structure, secondary-structure propensity, or the spatial arrangement of hydrophobic and charged residues, all of which can dominate hemolysis, solubility, and surface interactions. As a consequence, the controller can improve predicted scores but cannot yet explain those changes mechanistically. The current paper therefore avoids residue-level mechanism claims and treats property improvement as a model-output result.
 
\paragraph{Limited Controls.} The matched-start ES and aggressive analyses are intended as illustrative comparisons rather than fully controlled benchmarks. They draw on a single quartile-stratified sample of representative starting peptides under one controller configuration, so we report aggregate shifts rather than replicated statistical estimates. The baselines are also not matched on compute or edit depth (ES scores the full single-substitution neighborhood, while the aggressive setting is free to vary sequence length), and the comparison set is drawn from our own generator pool. A budget-controlled benchmark addressing these points is outlined in the Future Work subsection.
 
% ------------------------------------------------------------
\subsection{Future Work}
\label{sec:future_work}
 
\paragraph{Search Benchmarks.} A near-term priority is to convert the local-search headroom in Figure~\ref{fig:fixed_length_oracle} into an improved controller and a full benchmark. Algorithmically, the agent should adopt explicit accept--reject criteria on the composite score $S$, candidate buffering over the local neighborhood, and local-neighborhood proposals so that tool use cannot accept composite-score regressions when a better one-step neighbor is available; beyond single-step refinement, richer search procedures such as beam search, Monte Carlo tree search, or evolutionary strategies, combined with Pareto-based multi-objective ranking \cite{angelo2023multi}, would address candidates that are strong along one objective and only marginal along another, the methodological extension flagged at the end of the Methods section when introducing the uniformly weighted composite score. Evaluation should then compare conservative and aggressive Pepti-Agent variants with compute-matched sampled local search, ES, and length-matched random or permutation controls under the same starting pool, property predictors, and explicit evaluation budget. Because the aggressive setting improved matched-start scores while changing sequence length in most cases, the benchmark should separately control edit-policy breadth, local-neighborhood search, and prediction-call budget, so that the contribution of agentic orchestration can be quantified rather than asserted. The same benchmark protocol can also support a head-to-head comparison against ChemCrow \cite{bran2024chemtools}, SciToolAgent \cite{scitoolagent2025}, and DrugPilot \cite{drugpilot2025} on a shared peptide-developability task, which is currently outside the scope of this system's demonstration.
 
\paragraph{Structure-Aware Modeling.} The most important methodological extension is to incorporate structure-aware modeling into the agent loop. At present, candidate evaluation is based on sequence alone, even though hemolysis, solubility, and fouling behavior are strongly influenced by peptide conformation and spatial charge or hydrophobic patterning. A practical next step is therefore to couple the agent to a structure-prediction module, such as ESMFold \cite{esmfold2023} or AlphaFold-style methods \cite{jumper2021highly}, and extract structural descriptors that can supplement sequence-based predictors. In the same framework, mutation proposals could be guided by predicted structure rather than predictor scores alone, allowing the agent to target residues on hydrophobic faces, disordered termini, or aggregation-prone regions more deliberately.
 
\paragraph{Experimental Validation.} Because all results reported here are model-internal, the ultimate test of any candidate is the experimental measurement of hemolysis, solubility, and non-fouling behavior. Pepti-Agent is designed to make this interface straightforward rather than to perform the measurements itself: by exposing generation, prediction, mutation, and aggregation as MCP tools, the framework allows externally generated experimental data to be fed back as an updated training signal or as a new acceptance criterion without modifying the underlying model implementations. A natural use of the framework is therefore a computational and experimental collaboration in which top candidates from the pooled-refinement, ES, and aggressive-refinement analyses are synthesized and assayed by an experimental partner, and the resulting measurements are used to recalibrate the predictors and update the controller's decision logic in an iterative design--build--test cycle. A formal benchmark harness and automated test suite are natural deliverables that can be developed entirely computationally in advance of any such collaboration.
 
\section{Conclusion}
\label{sec:conclusion}
 
We introduced Pepti-Agent, an inspectable, MCP-orchestrated workflow that integrates peptide-specific generators, sequence-based property predictors, and two complementary single-site mutation operators behind a single tool interface. On a pool of 300 generated peptides, conservative Pepti-Agent refinement eliminated all remaining property-threshold failures while exposing a small but statistically significant cross-property trade-off in predicted non-fouling; on a quartile-stratified matched-start subset, the conservative controller was outscored in all 24 cases by exhaustive fixed-length single-substitution search, which is by construction the best single-edit neighbor; an aggressive edit policy exceeded both the conservative agent and ES on most starts, at the cost of changing sequence length in 21 of 24 starts. Consistent with the scope-limited claim made in the abstract, these results are model-internal: no candidate has been experimentally validated, the matched-start comparisons are search-space diagnostics rather than benchmarks, and the present manuscript does not claim that Pepti-Agent is a validated peptide-design optimizer or that it outperforms simpler search procedures. The contribution is a peptide-specific agentic workflow together with a concrete set of diagnostics that identify where adaptive tool use is currently strong (feasibility recovery) and where it is currently bounded (within-feasible composite-score optimization under fixed edit depth). Pepti-Agent provides a transparent workflow for predictor-guided peptide refinement and a reproducible basis for future benchmarking and experimental validation.
 
\begin{acknowledgement}
The authors thank the members of the Farimani Lab for helpful discussions and feedback. H.C. acknowledges support from the Carnegie Mellon University Biomedical Engineering Department.
\end{acknowledgement}

%%%%%%%%%%%%%%%%%%%%%%%%%%%%%%%%%%%%%%%%%%%%%%%%%%%%%%%%%%%%%%%%%%%%%
%% The same is true for Supporting Information, which should use the
%% suppinfo environment.
%%%%%%%%%%%%%%%%%%%%%%%%%%%%%%%%%%%%%%%%%%%%%%%%%%%%%%%%%%%%%%%%%%%%%
\begin{suppinfo}
 
The Supporting Information is available free of charge at \url{https://github.com/houxuc-rgb/AgentPeptide.git}. It contains Section S1, Codebase and Computing Environment; Section S2, Models and Checkpoints; Section S3, Prompts; Section S4, Detailed Methods; and Section S5, Full Results Tables and Source Data. These sections include repository structure, software and hardware environment, reproduction entry points, generator and predictor checkpoint identifiers, ESM2 fallback mutator details, LLM controller settings, MCP prompt resources, composite-score and threshold definitions, the 24-start stratified sampling rule, termination conditions, pooled 300-peptide source data, Wilcoxon and bootstrap statistics, ES results, aggressive-refinement results, figure source-data files, and the source-data manifest.
 
\end{suppinfo}

\bibliography{acs-achemso}
 
%%%%%%%%%%%%%%%%%%%%%%%%%%%%%%%%%%%%%%%%%%%%%%%%%%%%%%%%%%%%%%%%%%%%%
%% Supporting Information (merged; content unchanged)
%%%%%%%%%%%%%%%%%%%%%%%%%%%%%%%%%%%%%%%%%%%%%%%%%%%%%%%%%%%%%%%%%%%%%
\clearpage
\pagenumbering{arabic}
\renewcommand*{\thepage}{S\arabic{page}}
\Urlmuskip=0mu plus 1mu\relax
% turn on numbered sections for the SI only (inline form of \SectionNumbersOn,
% which is preamble-only in achemso); leaves the main paper headings unchanged
\makeatletter
\let\@startsection\acs@startsection
\makeatother
\setcounter{section}{0}
\setcounter{table}{0}
\setcounter{figure}{0}
\setcounter{equation}{0}
\renewcommand*{\thesection}{S\arabic{section}}
\renewcommand*{\thesubsection}{\thesection.\arabic{subsection}}
\renewcommand*{\thetable}{S\arabic{table}}
\renewcommand*{\thefigure}{S\arabic{figure}}
\renewcommand*{\theequation}{S\arabic{equation}}
 
{\centering\LARGE\bfseries Supporting Information for Pepti-Agent\par}
\vspace{1.5em}
\tableofcontents
 
\section{Codebase and Computing Environment}
 
\subsection{Repository Structure}
 
The code repository is available at \url{https://github.com/houxuc-rgb/AgentPeptide.git}.
 
\begin{table}[h]
\centering
\caption{Top-level codebase structure used by the MCP server.}
\begin{tabular}{ll}
\toprule
Path & Role \\
\midrule
\url{server.py} & MCP server entry point; registers tools and prompt resources. \\
\url{tools/} & PeptideGPT generation, PeptideBERT prediction, mutation, and result aggregation tools. \\
\url{prompts/} & MCP prompt resources exposed to clients. \\
\url{configs/} & PeptideBERT model construction code. \\
\url{requirements.txt} & Python package requirements for the MCP server and optional CLI mode. \\
\bottomrule
\end{tabular}
\end{table}
 
\subsection{Software and Hardware Environment}
 
\begin{table}[h]
\centering
\caption{Local computing environment used for code inspection and predictor execution.}
\begin{tabular}{ll}
\toprule
Item & Value \\
\midrule
Operating system & Ubuntu 24.04.4 LTS \\
CPU & Intel Core i7-8086K, 6 cores / 12 threads \\
System memory & 15 GiB RAM, 4.0 GiB swap \\
GPU & NVIDIA GeForce RTX 2080 Ti, 11264 MiB \\
NVIDIA driver & 580.126.09 \\
Python & 3.13.5 \\
PyTorch & 2.11.0+cu130 \\
CUDA reported by PyTorch & 13.0 \\
Transformers & 5.4.0 \\
fair-esm / esm & 2.0.0 \\
OpenAI Python package & 2.30.0 \\
NumPy & 2.4.3 \\
Pandas & 3.0.1 \\
SciPy & 1.17.1 \\
PyYAML & 6.0.3 \\
MCP Python package & Installed; package did not expose \code{\_\_version\_\_}. \\
\bottomrule
\end{tabular}
\end{table}
 
\subsection{Reproduction Entry Points}
 
The MCP server can be launched in three modes:
\begin{lstlisting}[style=cmd]
python server.py --transport stdio
python server.py --transport sse --port 8000
python server.py --transport cli
\end{lstlisting}
 
The optional CLI mode requires \code{OPENAI\_API\_KEY}. The MCP tools themselves do not require an LLM API key; local predictors and deterministic mutation tools are called by the MCP client.
 
\section{Models and Checkpoints}
 
\subsection{PeptideGPT Generator Checkpoints}
 
The three sequence generators are Hugging Face causal language model checkpoints loaded through \code{AutoModelForCausalLM.from\_pretrained}. Each generator uses stochastic sampling with \code{max\_new\_tokens = 50}, \code{do\_sample = True}, \code{top\_k = 50}, \code{top\_p = 0.95}, \code{temperature = 0.7}, and \code{num\_return\_sequences = 1}.
 
\begin{table}[h]
\centering
\caption{PeptideGPT generator identifiers.}
\begin{tabular}{lll}
\toprule
Property target & Code path & Hugging Face checkpoint \\
\midrule
Solubility & \url{tools/generate_soluability.py} & \url{aayush14/PeptideGPT_soluble} \\
Non-hemolytic & \url{tools/generate_non_hemolysis.py} & \url{aayush14/PeptideGPT_non_hemolytic} \\
Non-fouling & \url{tools/generate_non_fouling.py} & \url{aayush14/PeptideGPT_non_fouling} \\
\bottomrule
\end{tabular}
\end{table}
 
Generator seeds are short uppercase amino-acid strings. If no valid seed is supplied, the server draws a random two-residue seed from the standard amino-acid alphabet. The controller prompt recommends hydrophilic or charged residues for seeded generation: \code{K}, \code{R}, \code{E}, \code{D}, \code{S}, \code{T}, and \code{G} for non-hemolytic generation; \code{K}, \code{E}, \code{D}, and \code{S} for solubility; and \code{S}, \code{G}, \code{E}, and \code{K} for non-fouling generation.
 
\subsection{PeptideBERT-Style Predictor Checkpoints}
 
The predictor checkpoint root is set by \code{AGENTPMCP\_CHECKPOINTS\_DIR}; the run directories used here are listed in Table~\ref{tab:predictor_checkpoints}.
 
Each predictor checkpoint directory contains \code{config.yaml}, \code{network.py}, and \code{model.pt}. The configs set \code{epochs = 50}, \code{batch\_size = 32}, \code{vocab\_size = 25}, hidden size \code{480}, \code{12} hidden layers, \code{12} attention heads, dropout \code{0.15}, AdamW learning rate \code{1.0e-5}, and a ReduceLROnPlateau scheduler. All three inspected configs set \code{space\_tokens: true}.
 
\begin{table}[h]
\centering
\caption{PeptideBERT-style classifier head identifiers.}
\label{tab:predictor_checkpoints}
\begin{tabular}{llll}
\toprule
Property & Tool & Run directory & Weight file \\
\midrule
Solubility & \code{predict\_solubility} & \url{sol-0114_1359} & \url{model.pt} \\
Hemolysis & \code{predict\_hemolysis} & \url{hemo-0328_1451} & \url{model.pt} \\
Non-fouling & \code{predict\_non\_fouling} & \url{nf-0210_0959} & \url{model.pt} \\
\bottomrule
\end{tabular}
\end{table}
 
The loader defaults to \url{Rostlab/prot_bert} for tokenization and \url{Rostlab/prot_bert_bfd} for the ProtBERT backbone when those fields are absent from the checkpoint config. The local loader also resolves Hugging Face snapshots from \url{~/.cache/huggingface/hub} when a cached snapshot is available.
 
\subsection{ESM2 Fallback Mutator}
 
The fallback mutator in \url{tools/mutation.py} uses \code{esm.pretrained.esm2\_t6\_8M\_UR50D()}. It scans all positions and all standard amino-acid substitutions, excludes the wild-type residue at each position, computes the mutant minus wild-type log-likelihood difference, and returns the highest-scoring single substitution. No top-\(k\) truncation is applied.
 
\subsection{LLM Controller}
 
The interactive CLI controller uses an OpenAI-compatible Chat Completions API through the \code{openai} Python package. The default model ID is read from \code{AGENTPMCP\_LLM\_MODEL}; if unset, it defaults to \code{gpt-4o}.
 
For the manuscript experiments, \code{AGENTPMCP\_LLM\_MODEL} was set to Anthropic Claude Sonnet 4.6 (model ID \code{claude-sonnet-4-6}); the \code{gpt-4o} default was not used. All conservative and aggressive Pepti-Agent runs reported in the main text were executed against this model.
 
\section{Prompts}

\subsection{LLM-Guided Mutator and Controller Prompt}
 
The CLI controller prompt is embedded in \url{server.py}. The full system prompt text used by the controller is reproduced below, divided into its functional sections, with Unicode arrows and bullets normalized to ASCII for portability in the SI source.
 
\promptlabel{Role and core directive}
\begin{promptbox}
You are AgentP, an autonomous peptide engineering agent. You MUST
act by calling tools -- never just describe what you would do. Every
response from you MUST contain at least one tool call until you have a
final accepted candidate. Text-only replies that say "I will now ..."
without a tool call are FORBIDDEN.
\end{promptbox}
 
\promptlabel{Tool usage rules}
\begin{promptbox}
## TOOL USAGE RULES
- PeptideGPT generators expect a short amino acid seed (1-3 uppercase
  letters like "KR", "GS"), NOT natural language.
  - Non-hemolytic: hydrophilic seeds -- K, R, E, D, S, T, G
  - Soluble: charged / polar seeds -- K, E, D, S
  - Non-fouling: hydrophilic seeds -- S, G, E, K
- For property predictions, ALWAYS pass sequences as JSON with
  space_tokens: true, e.g.
  '{"sequences": "GSAEKRG", "space_tokens": true}'.
  Without space_tokens ProtBERT tokenizes incorrectly.
\end{promptbox}
 
\promptlabel{Scoring criteria}
\begin{promptbox}
## SCORING CRITERIA
- Hemolysis:   p_positive < 0.5 -> non-hemolytic (GOOD)
- Solubility:  p_positive >= 0.5 -> soluble        (GOOD)
- Non-fouling: p_positive >= 0.5 -> non-fouling     (GOOD)
A candidate PASSES when ALL three criteria are met simultaneously.
\end{promptbox}
 
\promptlabel{Iterative refinement protocol}
\begin{promptbox}
## ITERATIVE REFINEMENT PROTOCOL (follow exactly)
 
### Phase 0 -- Refine existing sequence (REFINE-EXISTING mode)
If the user provides a starting peptide sequence to refine:
1. SKIP all generators. Use the provided sequence as-is.
2. Immediately predict ALL THREE properties on the provided sequence.
3. Call evaluate_results to get a unified baseline summary.
4. Record the baseline scores. If it already passes all three -> report
   success and stop. Otherwise note which properties fail.
5. Proceed directly to Phase 2 (mutation refinement) using this sequence
   as the starting best candidate.
6. If mutation stalls (2 consecutive rounds with no improvement on the
   worst property), THEN and ONLY THEN use a generator targeting the
   worst property to get a fresh candidate, and continue Phase 2 with it.
   But always compare back against the user's original sequence scores.
 
### Phase 1 -- Seed generation (up to 3 candidates)
1. Call a generator (choose based on the user's priority, or default to
   generate_non_hemolytic_peptide).
2. Immediately predict ALL THREE properties on the generated sequence.
3. Call evaluate_results to get a unified summary.
4. Record the scores. If the candidate passes all three -> ACCEPT, go to
   Phase 3. Otherwise note which properties failed.
5. Repeat steps 1-4 up to 3 times with different seeds, picking the
   generator that targets the worst-failing property each time.
 
### Phase 2 -- Mutation refinement (up to 5 rounds per candidate)
Take the best candidate from Phase 1 (or Phase 0).
1. Decide which edit will most help the worst-failing property, then call
   mutate_sequence with:
     - sequence: current best sequence (no spaces)
     - operation: "substitute" | "insert" | "delete"
     - position: 0-indexed residue (for insert, position==len appends)
     - amino_acid: single letter ACDEFGHIKLMNPQRSTVWY (omit for delete)
   Heuristics:
     - To REDUCE hemolysis: substitute hydrophobic / cationic-amphipathic
       residues (L, I, V, F, W, K on hydrophobic face) with polar / neutral
       ones (S, T, G, N, Q).
     - To INCREASE solubility: substitute hydrophobic residues with charged
       / polar ones (K, R, E, D, S, T, N, Q).
     - To INCREASE non-fouling: substitute sticky / hydrophobic residues
       with hydrophilic / zwitterionic ones (S, G, E, K, D).
     - Use insert/delete sparingly -- only when length itself is the problem
       (e.g. trim a hydrophobic tail).
   If you cannot decide which residue to change, call esm2_mutate_sequence
   instead -- it picks the best single substitution automatically.
2. mutate_sequence returns JSON with mutated_sequence; use that field
   directly. esm2_mutate_sequence returns the raw mutated string.
3. Immediately predict ALL THREE properties on the mutated sequence.
4. Call evaluate_results.
5. Compare to the previous best:
   - If strictly better (more properties pass, or same passes with better
     scores on the failing property) -> adopt as new best.
   - If worse or no improvement -> discard, keep previous best.
6. If the candidate now passes all three -> ACCEPT, go to Phase 3.
7. If 2 consecutive mutations show no improvement on the worst property,
   STOP mutating. Instead go back to Phase 1 and generate a fresh
   candidate using the generator that targets the worst-failing property.
8. Repeat up to 5 mutation rounds per candidate, and up to 3 full
   generate-then-mutate cycles total.
 
### Phase 3 -- Final report
Print a summary with:
- Best sequence found
- All three property scores
- Pass/fail status for each
- How many generation + mutation steps were used
\end{promptbox}
 
\promptlabel{Critical behavioral rules}
\begin{promptbox}
## CRITICAL BEHAVIORAL RULES
- NEVER stop to narrate without calling a tool. If you want to explain
  your reasoning, include the explanation AND a tool call in the same
  response.
- After every mutation call (mutate_sequence or esm2_mutate_sequence),
  you MUST immediately call the three predict_* tools. Never skip
  prediction after mutation.
- After every round of three predictions, you MUST call evaluate_results.
- Keep a running "best candidate" with its scores. Always compare new
  candidates against it.
- If you are unsure what to do next, call the generator targeting the
  worst-failing property.
- For mutation tools, pass the sequence WITHOUT spaces (e.g. "GSKWQ",
  not "G S K W Q"). Spaces are for PeptideBERT only.
- PREFER mutate_sequence -- it lets you pick the operation, position, and
  residue based on the property scores. Fall back to esm2_mutate_sequence
  only when you cannot decide which residue to change.
- mutate_sequence returns JSON; parse and use the "mutated_sequence" field
  for the next prediction step.
- If a tool call returns an error, DO NOT give up. Try again with
  corrected input, or switch strategy (e.g. regenerate instead of
  mutate). Errors are expected -- recover and continue.
- NEVER end with a "could not complete" message if you still have
  generation or mutation attempts remaining.
\end{promptbox}
 
For an explicit \code{refine SEQUENCE [goals]} command, the CLI builds the following user message template:
\promptlabel{Refine command user template}
\begin{userpromptbox}
I have an existing peptide sequence that I want you to refine and improve.
 
Starting sequence: <RAW_SEQUENCE>
Optimization goals: <GOALS OR all three properties>
 
Follow the REFINE-EXISTING protocol: predict its current properties first,
then iteratively mutate to improve the failing ones. Do NOT generate a new
sequence from scratch -- start from this exact sequence.
\end{userpromptbox}
 
For a raw peptide sequence without the \code{refine} prefix, the same template is used with optimization goals set to all three properties: soluble, non-hemolytic and non-fouling.
 
\subsection{MCP Prompt Resource: peptide\_generation}
 
\begin{genpromptbox}
You are AgentP, a methodical AI peptide engineer.
Your objective is to generate peptides using the correct generation tool.
 
For multi-property design or iterative search, prefer optimize_peptide.
For single-property generation, pick EXACTLY ONE generator based on the user's intent:
- soluble / solubility / water-soluble              -> generate_soluble_peptide
- non-hemolytic / hemolysis / RBC-safe              -> generate_non_hemolytic_peptide
- non-fouling / anti-fouling / biofouling-resistant -> generate_non_fouling_peptide
 
If optimize_peptide is not available and multiple properties are mentioned, prioritise:
  1) non-hemolytic  2) non-fouling  3) soluble
 
Before using a generator, briefly state which tool you chose and the property it targets.
Do not ask the user for a seed unless they explicitly want to provide one.
If no seed is supplied, choose a short uppercase 1-3 residue seed yourself.
After generation, report the final peptide sequence clearly.
\end{genpromptbox}
 
The inspected MCP server does not register an \code{optimize\_peptide} tool; when that high-level workflow is unavailable, the prompt instructs clients to drive the exposed low-level generation, prediction, mutation, and evaluation tools directly.
 
\subsection{MCP Prompt Resource: property\_prediction}
 
\begin{predpromptbox}
You are AgentP, a peptide evaluation assistant.
 
For any given peptide sequence, run ALL THREE property predictors:
  1. predict_hemolysis
  2. predict_solubility
  3. predict_non_fouling
 
Then call evaluate_results with the three JSON outputs to produce a unified summary.
 
Use the returned p_positive values directly:
  - Soluble       when solubility p_positive >= 0.5
  - Non-hemolytic when hemolysis p_positive <= 0.5
  - Non-fouling   when non-fouling p_positive >= 0.5
 
Report the raw probabilities explicitly and state whether each threshold passes.
 
Do not skip any predictor.
Keep the final summary concise and explicitly list the three outcomes.
\end{predpromptbox}
 
\subsection{MCP Prompt Resource: iterative\_refinement}
 
\begin{refpromptbox}
You are AgentP, a peptide optimization controller.
 
Goal: design a peptide that is SOLUBLE and NON-FOULING, with LOW HEMOLYSIS.
 
Preferred workflow:
  - Call optimize_peptide for the full closed-loop search whenever possible.
  - If you must drive the low-level tools manually, do it as a strict loop:
    1. Evaluate the baseline with all three predictors.
    2. Mutate the current best sequence.
    3. Re-evaluate the mutant with all three predictors.
    4. Compare the new probabilities against the current best.
    5. If progress stalls, regenerate toward the weakest property and continue.
    6. Stop only after at least one full mutation + re-evaluation cycle.
 
Numeric objective:
  - Maximise solubility p_positive
  - Maximise non-fouling p_positive
  - Minimise hemolysis p_positive
 
Thresholds:
  - Soluble       when solubility p_positive >= 0.5
  - Non-fouling   when non-fouling p_positive >= 0.5
  - Non-hemolytic when hemolysis p_positive <= 0.5
 
Execution style:
  - Be decisive and call tools when the next step is clear.
  - Keep reasoning concise; focus on numeric comparisons and the next action.
  - Preserve the best sequence seen so far across all rounds.
 
Final answer checklist (mandatory):
  - original_sequence: <seq>
  - modified_sequence: <seq>
  - reevaluation_done: YES
  - baseline_scores: {solubility_p_positive: <v>, non_fouling_p_positive: <v>, hemolysis_p_positive: <v>}
- modified_scores:  {solubility_p_positive: <v>, non_fouling_p_positive: <v>, hemolysis_p_positive: <v>}
- decision: improved / not improved -- with numeric justification
\end{refpromptbox}
 
\section{Detailed Methods}
 
\subsection{Composite Score and Thresholds}
 
The composite score used throughout the refinement analyses is
\begin{equation}
\Sscore = p_{\mathrm{sol}} + p_{\mathrm{nf}} + (1 - p_{\mathrm{hem}}).
\end{equation}
 
The classifier-pass thresholds are \(p_{\mathrm{sol}} \ge 0.5\), \(p_{\mathrm{nf}} \ge 0.5\), and \(p_{\mathrm{hem}} \le 0.5\). The stricter design-pass thresholds listed in the processing checklist are \(p_{\mathrm{sol}} \ge 0.7\), \(p_{\mathrm{nf}} \ge 0.6\), and \(p_{\mathrm{hem}} \le 0.4\). The threshold-processing checklist is archived at \path{results/customed/data/custom_processing_config.csv}.
 
\subsection{Conservative and Aggressive Controller Settings}
 
\begin{table}[h]
\centering
\caption{Controller configuration values used for the conservative and aggressive refinement settings.}
\begin{tabular}{@{}L{0.24\textwidth}L{0.34\textwidth}L{0.34\textwidth}@{}}
\toprule
Setting & Conservative controller & Aggressive controller \\
\midrule
Classifier pass thresholds & 0.5 / 0.5 / 0.5 & 0.5 / 0.5 / 0.5 \\
Round budget & 5 mutation rounds in prompt & 8 rounds in the aggressive refinement setting \\
Insertion/deletion allowed & Sparingly, prompt-guided & Yes \\
Multi-site edits per round & No, single deterministic edit & No, one proposed edit per step \\
\bottomrule
\end{tabular}
\end{table}
 
For the aggressive controller, edits are prioritized by target property. Hemolysis reduction first substitutes hydrophobic residues with \code{P}, \code{G}, \code{S}, \code{D}, \code{E}, \code{T}, or \code{N}, then substitutes cationic residues with neutral or anionic residues, then inserts \code{P}/\code{G}, then deletes hydrophobic or cationic residues. Solubility improvement substitutes hydrophobic residues with charged or polar residues, permits terminal charged insertions, and can delete hydrophobic residues. Non-fouling improvement substitutes hydrophobic or cationic residues with \code{S}, \code{G}, \code{T}, \code{N}, or \code{Q}, and permits \code{S}/\code{G} terminal insertions. The staged aggressive output summaries are archived in:
\begin{itemize}
\item \path{nonagent/aggressive_refinement_summary.csv}
\item \path{nonagent/agg/aggressive_comparison_matched_records.csv}
\end{itemize}
 
\subsection{24-Start Stratified Sampling Rule}
 
The 24-start comparison set was verified at:
\begin{lstlisting}[style=path]
nonagent/selected_peptides_for_optimizer_comparison.csv
\end{lstlisting}
 
The selection rule recorded in the file is: two representative peptides from each source-pool score quartile, ranked by baseline \(\Sscore\). This gives \(3\) source pools \(\times\) \(4\) quartiles \(\times\) \(2\) starts \(=24\) starts. The script \url{generate_variants.py} preserves CSV row order and does not implement an additional tie-breaker; therefore any tied ordering must come from the upstream CSV construction.
 
\begin{table}[h]
\centering
\caption{Twenty-four starting sequences used for ES and aggressive comparisons.}
\resizebox{\textwidth}{!}{%
\begin{tabular}{lllllrrrr}
\toprule
Start ID & Pool & Quartile & Sequence & Length & \(p_{\mathrm{sol}}\) & \(p_{\mathrm{nf}}\) & \(p_{\mathrm{hem}}\) & \(\Sscore\) \\
\midrule
sol\_Q1\_low\_1 & sol & Q1\_low & KKEAA & 5 & 0.672515 & 0.631395 & 0.394007 & 1.909903 \\
sol\_Q1\_low\_2 & sol & Q1\_low & KSEIK & 5 & 0.704873 & 0.644515 & 0.412146 & 1.937242 \\
sol\_Q2\_mid\_low\_3 & sol & Q2\_mid\_low & DEDNE & 5 & 0.701690 & 0.677965 & 0.393765 & 1.985890 \\
sol\_Q2\_mid\_low\_4 & sol & Q2\_mid\_low & EKSRN & 5 & 0.725184 & 0.666588 & 0.393815 & 1.997958 \\
sol\_Q3\_mid\_high\_5 & sol & Q3\_mid\_high & DSDKE & 5 & 0.724031 & 0.689090 & 0.400876 & 2.012246 \\
sol\_Q3\_mid\_high\_6 & sol & Q3\_mid\_high & SKKTD & 5 & 0.728873 & 0.681304 & 0.392104 & 2.018073 \\
sol\_Q4\_high\_7 & sol & Q4\_high & KKEQEDEENE & 10 & 0.712324 & 0.697849 & 0.376963 & 2.033209 \\
sol\_Q4\_high\_8 & sol & Q4\_high & DESNDTN & 7 & 0.728299 & 0.678546 & 0.368680 & 2.038164 \\
nf\_Q1\_low\_9 & nf & Q1\_low & KKEKR & 5 & 0.710246 & 0.689734 & 0.468284 & 1.931696 \\
nf\_Q1\_low\_10 & nf & Q1\_low & SSERPVCPQC & 10 & 0.728914 & 0.619689 & 0.379485 & 1.969119 \\
nf\_Q2\_mid\_low\_11 & nf & Q2\_mid\_low & KGKAWKR & 7 & 0.723540 & 0.643782 & 0.368943 & 1.998379 \\
nf\_Q2\_mid\_low\_12 & nf & Q2\_mid\_low & GSKAGE & 6 & 0.724450 & 0.645257 & 0.363428 & 2.006279 \\
nf\_Q3\_mid\_high\_13 & nf & Q3\_mid\_high & KSSST & 5 & 0.729508 & 0.670775 & 0.378617 & 2.021666 \\
nf\_Q3\_mid\_high\_14 & nf & Q3\_mid\_high & GGGTRGGGA & 9 & 0.720497 & 0.667905 & 0.358290 & 2.030111 \\
nf\_Q4\_high\_15 & nf & Q4\_high & SKGDPIS & 7 & 0.729712 & 0.671146 & 0.359786 & 2.041072 \\
nf\_Q4\_high\_16 & nf & Q4\_high & GGSPS & 5 & 0.727655 & 0.689575 & 0.366946 & 2.050283 \\
hem\_Q1\_low\_17 & hem & Q1\_low & RRSRR & 5 & 0.701267 & 0.640494 & 0.485521 & 1.856240 \\
hem\_Q1\_low\_18 & hem & Q1\_low & KKSFF & 5 & 0.706136 & 0.570694 & 0.395801 & 1.881030 \\
hem\_Q2\_mid\_low\_19 & hem & Q2\_mid\_low & KRDKR & 5 & 0.714772 & 0.668836 & 0.461214 & 1.922395 \\
hem\_Q2\_mid\_low\_20 & hem & Q2\_mid\_low & RDEKW & 5 & 0.716595 & 0.625862 & 0.411867 & 1.930590 \\
hem\_Q3\_mid\_high\_21 & hem & Q3\_mid\_high & KSERLKQLEK & 10 & 0.722107 & 0.613722 & 0.381677 & 1.954153 \\
hem\_Q3\_mid\_high\_22 & hem & Q3\_mid\_high & KSKWRAR & 7 & 0.724891 & 0.614971 & 0.367015 & 1.972848 \\
hem\_Q4\_high\_23 & hem & Q4\_high & TSGKV & 5 & 0.729601 & 0.651775 & 0.383220 & 1.998157 \\
hem\_Q4\_high\_24 & hem & Q4\_high & STSKWKS & 7 & 0.729671 & 0.647946 & 0.369511 & 2.008107 \\
\bottomrule
\end{tabular}}
\end{table}
 
\subsection{Termination Conditions}
 
The CLI controller prompt permits up to five mutation rounds per candidate and up to three initial generated candidates. The CLI driver has a hard cap of 50 model turns and a stall detector that nudges text-only responses up to three times. The aggressive controller uses \code{MAX\_ROUNDS = 8}. The inspected code does not impose a wall-clock limit.
 
\section{Full Results Tables and Source Data}
 
\subsection{Pooled 300-Peptide Data}
 
The checklist identifies the following source files as ready for SI inclusion:
\begin{lstlisting}[style=path]
results/origin/sampled_{hem,nf,sol}.json
results/refined/sampled_{hem,nf,sol}.json
results/figure_data/refinement_pair_records.csv
results/newKind/result/source_data_peptide_refinement_v6.csv
results/figure_data/refinement_pair_stats.csv
results/newKind/result/score_summary_v6.csv
results/newKind/result/quality_gate_summary_v6.csv
results/newKind/result/gate_transition_v6.csv
results/newKind/result/gate_sensitivity_v6.csv
\end{lstlisting}
 
These files are staged in repository-relative paths so that they can be archived with the manuscript source.
 
\subsection{Wilcoxon and Bootstrap Statistics}
 
Table~\ref{tab:wilcoxon} reports, for each of the four paired endpoints used in Figure~2a of the main text, the sample size \(n\), the Wilcoxon signed-rank statistic \(W\) (taken as the rank sum of positive paired differences, \(\sum r_+\)), the uncorrected two-sided \(p\)-value, the Bonferroni-corrected \(p\)-value (factor 4, one correction per endpoint), the paired effect size Cohen's \(d_z\), and the 95\% bootstrap confidence interval for \(d_z\). All endpoints are scored in the direction in which higher is preferred: predicted solubility, the non-hemolytic score \(1-p_{\mathrm{hem}}\), predicted non-fouling, and the composite developability score \(S\).
 
Paired tests were run with \code{scipy.stats.wilcoxon} using the default two-sided alternative; zero-difference pairs were handled with the \code{wilcox} zero-method, which discards exact ties (19 of 300 paired sequences were unchanged after refinement). Each 95\% confidence interval for \(d_z\) was obtained by 10{,}000 nonparametric bootstrap resamples of the paired difference vector, drawn with NumPy's \code{numpy.random.default\_rng} (PCG64) generator at a fixed seed of 11; the corresponding mean paired-\(\Delta\) intervals reported elsewhere in the SI and on the panel A x-axis use seed 7. The generating script is \code{results/newKind/make\_peptide\_refinement\_figure\_v6.py}, and the underlying paired score arrays are \code{results/origin/sampled\_\{hem,nf,sol\}.json} versus \code{results/refined/sampled\_\{hem,nf,sol\}.json}. The full per-endpoint summary, including means, medians, improved/unchanged/worsened counts, and the one-sided \(W\) and \(p\) values, is archived in \code{results/newKind/result/score\_summary\_v6.csv}.
 
\begin{table}[h]
\centering
\caption{Paired Wilcoxon signed-rank tests with Bonferroni correction across four endpoints, and paired Cohen's \(d_z\) with 95\% bootstrap confidence intervals (10{,}000 resamples; \code{numpy.random.default\_rng}, seed 11). \(W\) is the rank sum of positive paired differences (\(\sum r_+\)). Endpoints are scored in the preferred direction. Source: \code{results/newKind/result/score\_summary\_v6.csv}.}
\label{tab:wilcoxon}
\begin{tabular}{lrrrrrr}
\toprule
Endpoint & \(n\) & \(W\) & raw \(p\) & Bonferroni \(p\) & \(d_z\) & 95\% CI \\
\midrule
Solubility            & 300 & 36166 & \(3.73\times10^{-33}\) & \(1.49\times10^{-32}\) & \(+0.553\) & \([+0.488,\,+0.629]\) \\
\(1-p_{\mathrm{hem}}\) & 300 & 34070 & \(1.34\times10^{-25}\) & \(5.35\times10^{-25}\) & \(+0.497\) & \([+0.408,\,+0.602]\) \\
Non-fouling           & 300 & 10796 & \(3.80\times10^{-11}\) & \(1.52\times10^{-10}\) & \(-0.389\) & \([-0.535,\,-0.263]\) \\
Composite \(S\)       & 300 & 17566 & \(9.97\times10^{-2}\)  & \(3.99\times10^{-1}\)  & \(+0.098\) & \([-0.018,\,+0.193]\) \\
\bottomrule
\end{tabular}
\end{table}
 
\subsection{Fixed-Length ES Results}
 
The fixed-length ES results are available at:
\begin{lstlisting}[style=path]
nonagent/nonagent_greedy_summary.csv
\end{lstlisting}
 
The ES enumerates all single-position, same-length substitutions over the standard amino-acid alphabet, excludes the original residue, evaluates unique variants, and selects the best-scoring single substitution for each start. The variant-generation script is:
\begin{lstlisting}[style=path]
nonagent/generate_variants.py
\end{lstlisting}
 
The complete CSV includes original sequence, selected sequence, all three original probabilities, original score, all three selected probabilities, and selected score for all 24 starts.
 
\subsection{Aggressive Refinement Results}
 
The aggressive refinement summary is available at:
\begin{lstlisting}[style=path]
nonagent/aggressive_refinement_summary.csv
\end{lstlisting}
 
The aggressive figure and review-revision analysis scripts are:
\begin{lstlisting}[style=path]
nonagent/agg/make_aggressive_refinement_figure.py
nonagent/review_revisions/make_nonagent_review_revisions_v3.py
\end{lstlisting}
 
The aggressive source-data files contain original and final sequences, all three property scores, composite scores, length changes, and edit-policy summaries used in the corresponding main-text panel. The length-change and edit-distance summaries are archived at \code{nonagent/review\_revisions/aggressive\_edit\_policy\_breadth\_v3.csv} and \code{nonagent/review\_revisions/aggressive\_length\_stratified\_deltaS\_v3.csv}.
 
\subsection{Figure Source Data}
 
\begin{table}[h]
\centering
\caption{Source data files staged with the SI.}
\begin{tabular}{ll}
\toprule
Manuscript item & Source data \\
\midrule
Figure 2 refinement panels & \url{results/figure_data/refinement_pool_records.csv}; \url{results/figure_data/refinement_pool_stats.csv} \\
Figure 3 agent-vs-ES panels & \url{nonagent/try_figures/matched_agent_nonagent_source.csv}; \url{nonagent/review_revisions/agent_vs_nonagent_*_v3.csv} \\
Figure 4 aggressive panels & \url{nonagent/agg/aggressive_*}; \url{nonagent/review_revisions/aggressive_*_v3.csv} \\
Pooled 300 SI table & \url{results/newKind/result/source_data_peptide_refinement_v6.csv} \\
Quality gate sensitivity & \url{results/newKind/result/gate_sensitivity_v6.csv} \\
\bottomrule
\end{tabular}
\end{table}

\end{document}